\pdfoutput=1
\documentclass[runningheads]{llncs}
\usepackage{graphicx}
\usepackage{comment}
\usepackage{amsmath,amssymb} 
\usepackage{color}
\usepackage{url}
\usepackage{subfiles}
\usepackage{times}
\usepackage{epsfig}
\usepackage{graphicx}
\usepackage{amsmath}
\usepackage{amssymb}
\usepackage{gensymb}
\usepackage{caption}

\usepackage{booktabs}
\usepackage{multirow}
\usepackage{amsfonts}
\usepackage{threeparttable}
\usepackage{multirow}
\usepackage[toc,page]{appendix}

\usepackage{xspace}

\makeatletter
\DeclareRobustCommand\onedot{\futurelet\@let@token\@onedot}
\def\@onedot{\ifx\@let@token.\else.\null\fi\xspace}

\def\eg{\emph{e.g}\onedot} 
\def\ie{\emph{i.e}\onedot} 
 
\def\etc{\emph{etc}\onedot} 
 
\def\etal{\emph{et al}\onedot}
\makeatother

\begin{document}
\pagestyle{headings}
\mainmatter
\def\ECCVSubNumber{4549}  

\title{P$^{2}$Net: Patch-match and Plane-regularization \\ for Unsupervised Indoor Depth Estimation}
\titlerunning{P$^{2}$Net for Unsupervised Indoor Depth}
\authorrunning{Yu et al.}
\author{Zehao Yu$^*$\inst{1,2} \and
    Lei Jin$^*$\inst{1,2} \and
    Shenghua Gao$^{\dag}$\inst{1} \\
}
\institute{ShanghaiTech Univsertiy \and DGene Inc \\
    {\email \{yuzh,jinlei,gaoshh\}@shanghaitech.edu.cn} \\
    \small \url{https://github.com/svip-lab/Indoor-SfMLearner}
}

\maketitle

\begin{abstract}
This paper tackles the unsupervised depth estimation task in indoor environments. The task is extremely challenging because of the vast areas of non-texture regions in these scenes.
These areas could overwhelm the optimization process in the commonly used unsupervised depth estimation framework proposed for outdoor environments. 
However, even when those regions are masked out, the performance is still unsatisfactory.  
In this paper, we argue that the poor performance suffers from the non-discriminative point-based matching. To this end, we propose P$^2$Net. We first extract points with large local gradients and adopt patches centered at each point as its representation. Multiview consistency loss is then defined over patches. This operation significantly improves the robustness of the network training. 
Furthermore, because those textureless regions in indoor scenes (\eg, wall, floor, roof, \etc) usually correspond to planar regions, 
we propose to leverage superpixels as a plane prior. We enforce the predicted depth to be well fitted by a plane within each superpixel.
Extensive experiments on NYUv2 and ScanNet show that our P$^2$Net outperforms existing approaches by a large margin.

\keywords{Unsupervised Depth estimation, Patch-based Representation, Multiview Photometric Consistency, Piece-wise Planar Loss}
\newcommand\blfootnote[1]{%
    \begingroup
    \renewcommand\thefootnote{}\footnote{#1}%
    \addtocounter{footnote}{-1}%
    \endgroup
}
\blfootnote{$*$~Equal Contribution}
\blfootnote{$\dag$~Corresponding author}
\end{abstract}

\section{Introduction}
\label{sec:intro}

Depth estimation, as a fundamental problem in computer vision, bridges the gap between 2D images and 3D world. Lots of supervised depth estimation methods~\cite{eigen2014depth,fu2018deep,liu2015learning} have been proposed with the recent trend in convolution neural networks (CNNs). However, capturing a large number of images in different scenes with accurate ground truth depth requires expensive hardware and time~\cite{dai2017scannet,kitti,saxena2008make3d,nyu,diode_dataset}.
To overcome the above challenges, another line of work \cite{garg2016unsupervised,godard2017unsupervised,xie2016deep3d,zhou2017unsupervised} focuses on unsupervised depth estimation that only uses either stereo videos or monocular videos as training data. The key supervisory signal in these work is the appearance consistency between the real view and the view synthesized based on the estimated scene geometry and ego-motion of the camera. Bilinear interpolation~\cite{jaderberg2015spatial} based warping operation allows the 
training process to be fully differentiable.

While recent works of unsupervised depth estimation~\cite{yin2018geonet,Zhou_2019_ICCV,zou2018df} have demonstrated impressive results on outdoor datasets, 
the same training process may easily collapse~\cite{zhou2019moving} on indoor datasets such as NYUv2~\cite{nyu} or ScanNet~\cite{dai2017scannet}. The primary reason is that indoor environments contain large non-texture regions where the photometric consistency (the main supervisory signal in unsupervised learning) is unreliable. 
In such regions, the predicted depth might decay to infinite, while the synthesized view still has a low photometric error. 
Similar problems~\cite{godard2017unsupervised,godard2019digging,luo2018every,yin2018geonet} are also observed on outdoor datasets, especially in road regions. While the propotion of such regions is small on outdoor datasets, which would only lead to degradation in performance, the large non-texture regions on indoor scenarios can easily overwhelm the whole training process.

An intuitive try would be to mask out all the non-texture regions during the loss calculation. However, as the experimental results will demonstrate, merely ignoring the gradients from these non-texture regions still leads to inferior results. 
The reason is that we are minimizing per pixel (point) based multi-view photometric consistency error in the training process, where each point should be matched correctly across different views. 
Such point-based representation is not discriminative enough for matching in indoor scenes, since many other pixels in images could have the same intensity values. This operation could easily result in false matching. 
Even replacing bilinear sampling operation with the recent proposed linearized multi-sampling~\cite{jiang2019linearized} that creates a linear approximation with more samples in view synthesis still could not resolve the inherited deficiency in the discriminative representation of the point-based representation.
Instead, taking inspiration from traditional multi-view stereo approaches~\cite{furukawa2009accurate,schoenberger2016mvs} that represent a point with a local patch, we propose to replace point-based representation with a patch-based representation to increase the discriminative ability in the matching process.
Specifically, points with large local gradients are selected as our keypoints. We assume the same depth for pixels within a local window around every keypoint. We then project these local patches to different views with the predicted depth map and camera motion, and minimize multi-view photometric consistency error over the patches. Compared to point-based representation, our patch-based solution leads to a more distinctive characterization that produces more representative gradients with a wider basin of convergence.

Finally, to handle the rest large non-texture regions in indoor scenes, we draw inspiration from the recent success of work~\cite{furukawa2009manhattan,liu2018planenet,yu2019single} that leverages the plane prior for indoor scene reconstruction. We make the assumption that homogeneous-colored regions, for example, walls, can be approximated with a plane. Here we adopt a similar strategy with the previous work~\cite{concha2014using,concha2015dpptam} that approximates the planar regions with superpixels. Specifically, we first extract planar regions by superpixels~\cite{felzenszwalb2004efficient}, then use a planar consistency loss to enforce the predicted depth in these regions can be well fitted by a plane, \ie, low plane-fitting error within each superpixel. This allows our network to produce a more robust result.

Compared with MovingIndoor~\cite{zhou2019moving}, a pioneer work on unsupervised indoor depth estimation that requires to first establish sparse correspondences between consecutive frames,
and then propagates the sparse flows to the entire image,
our P$^2$Net is direct, and no pre-matching process is required. Therefore, there is no concern for falsely matched pairs that might misguide the training of the network. Further, the supervisory signal of MovingIndoor~\cite{zhou2019moving} comes from the consistency between the synthesized optical flow and the predicted flow of the network. Such indirect supervision might also lead to a sub-optimal result. Our P$^2$Net instead supervises the network from two aspects: local patches for textured regions and planar consistency for the non-texture regions. 

Our contributions can be summarized as follows: i) we propose to extract discriminative keypoints with large local gradients and use patches centered at each point as its representation. ii) patch-match: A patch-based warping process that assumes the same depth for pixels within a local patch is proposed for a more robust matching.  
iii) plane-regularization: we propose to use superpixels to represent those homogeneous-texture or non-texture piece-wise planar regions and regularize the depth consistency within each superpixel. 
On the one hand, our P$^2$Net leverages the discriminative patch-based representation that improves the matching robustness. On the other hand, our P$^2$Net encodes the piece-wise planar prior into the network. Consequently, our approach is more suitable for indoor scene depth estimation.
Extensive experiments on widely-used indoor datasets NYUv2~\cite{nyu} and ScanNet~\cite{dai2017scannet} demonstrate that P$^2$Net outperforms state-of-the-art by a large margin.

\section{Related Work}
\label{sec:relatedwork}
\subsection{Supervised Depth Estimation}
A vast amount of research has been done in the field of supervised depth estimation. With the recent trend in convolution neural networks (CNNs), many different deep learning based approaches have been proposed. Most of them frame the problem as a per-pixel regression problem. Particularly, Eigen et al.~\cite{eigen2014prediction} propose a multi-scale approach that predicts a coarse global depth maps based on the entire image and then refine the prediction with CNNs. Laina et al.~\cite{laina2016deeper} improve the performance of depth estimation by introducing a fully convolutional architecture with several up-convolution blocks. Kim et al.~\cite{crf1} use conditional random fields to refine the depth prediction. Recently, Fu et al.~\cite{fu2018deep} treat the problem from an ordinal regression perspective. With a carefully designed discretization strategy and an ordinal loss, their method is able to achieve new state-of-the-art results in supervised depth estimation. Other work focuses on combining depth estimation with other tasks, for example, semantic segmentation~\cite{jiao2018look,zhang2018joint} and surface norm estimation~\cite{eigen2015predicting,qi2018geonet}. Yin et al.~\cite{yin2019enforcing} show that high-order 3D geometric constraints, the so-called virtual normal, can further improve depth prediction accuracy. However, all of these methods rely on vast amounts of labeled data, which is still a large cost in both hardware and time.

\subsection{Unsupervised Depth Estimation}
Unsupervised learning of depth estimation has been proposed to ease the demand for large-scale labeled training data. One line of work exploits stereo images or videos~\cite{xie2016deep3d,garg2016unsupervised,godard2017unsupervised} as training data and trains a network to minimize the photometric error between synthesized view and real view. Godard et al.~\cite{godard2017unsupervised} introduce a left-right disparity consistency as regularization. Another line of work learns depth from monocular video sequences. Zhou et al.~\cite{zhou2017unsupervised} introduce a separate network to predict camera motion between input images. Their method learns to estimate depth and ego-motion simultaneously.
Later work also focuses on joint-learning by minimizing optical flow errors~\cite{ren2017unsupervised,yin2018geonet}, or combining SLAM pipelines into deep networks~\cite{shi2019self,wang2018learning}. However, none of the above approaches produce satisfactory results on indoor datasets. MovingIndoor~\cite{zhou2019moving} is the first work to study unsupervised depth estimation in indoor scenes. The authors propose an optical flow estimation network, SFNet, initialized with sparse flows from matching results of SURF~\cite{bay2006surf}. During training, the sparse flows are propagated iteratively from texture regions to non-texture regions and transformed into dense flows. The dense optical flows are used as the supervisory signal for the learning of the depth and pose. By contrast, we propose to supervise the training with a more discriminative patch-based multi-view photometric consistency error and regularize the depth within homogeneous-color regions with a planar consistency loss. Our method is direct, and no pre-matching process is required. Therefore, there is no concern for falsely matched pairs that might misguide the training of the network.

\subsection{Piece-wise Planar Scene Reconstruction}
Piece-wise planar reconstruction is an active research
topic in multi-view 3D reconstruction~\cite{furukawa2009manhattan,gallup2010piecewise}, SLAM~\cite{concha2014using,concha2015dpptam} and has drawn increasing attention recently~\cite{liu2018planenet,Yang_2018_ECCV,yu2019single,liu2019planercnn}. Traditional methods~\cite{furukawa2009accurate,gallup2010piecewise} generate plane hypotheses by fitting planes to triangulated 3D points, then assign hypotheses to each pixel via a global optimization. Concha and Civera~\cite{concha2014using,concha2015dpptam} used superpixels~\cite{felzenszwalb2004efficient} to describe non-texture region  in a monocular dense SLAM system. Their method has shown impressive reconstruction results. 
Raposo et al.~\cite{raposo2016pi} proposed $\pi$Match, a vSLAM pipeline with plane features to for a piecewise planar reconstruction. In their more recent work~\cite{8006257}, they recovered structure and motion from planar regions and combined these estimations into stereo algorithms.
Together with Deep CNNs, Liu et al.~\cite{liu2018planenet} learn to infer plane parameters and associates each pixel to a plane in a supervised manner. Yang and Zhou~\cite{Yang_2018_ECCV} learn a similar network with only depth supervision. Following work~\cite{liu2019planercnn,yu2019single} further formulate the planar reconstruction problem as an instance segmentation problem and have shown significant improvements. 
 Inspired by these work, we incorporate the planar prior for homogeneous-color regions into our unsupervised framework and propose a planar consistency loss to regularize the depth map in such regions in the training phrase. 

\section{Method}
\newcommand{\bA}{\mathbf{A}}
\newcommand{\bE}{\mathbf{E}}
\newcommand{\bX}{\mathbf{X}}
\newcommand{\bY}{\mathbf{Y}}

\label{sec:methods}
\begin{figure}[!t]
\centering
\includegraphics[width=12cm]{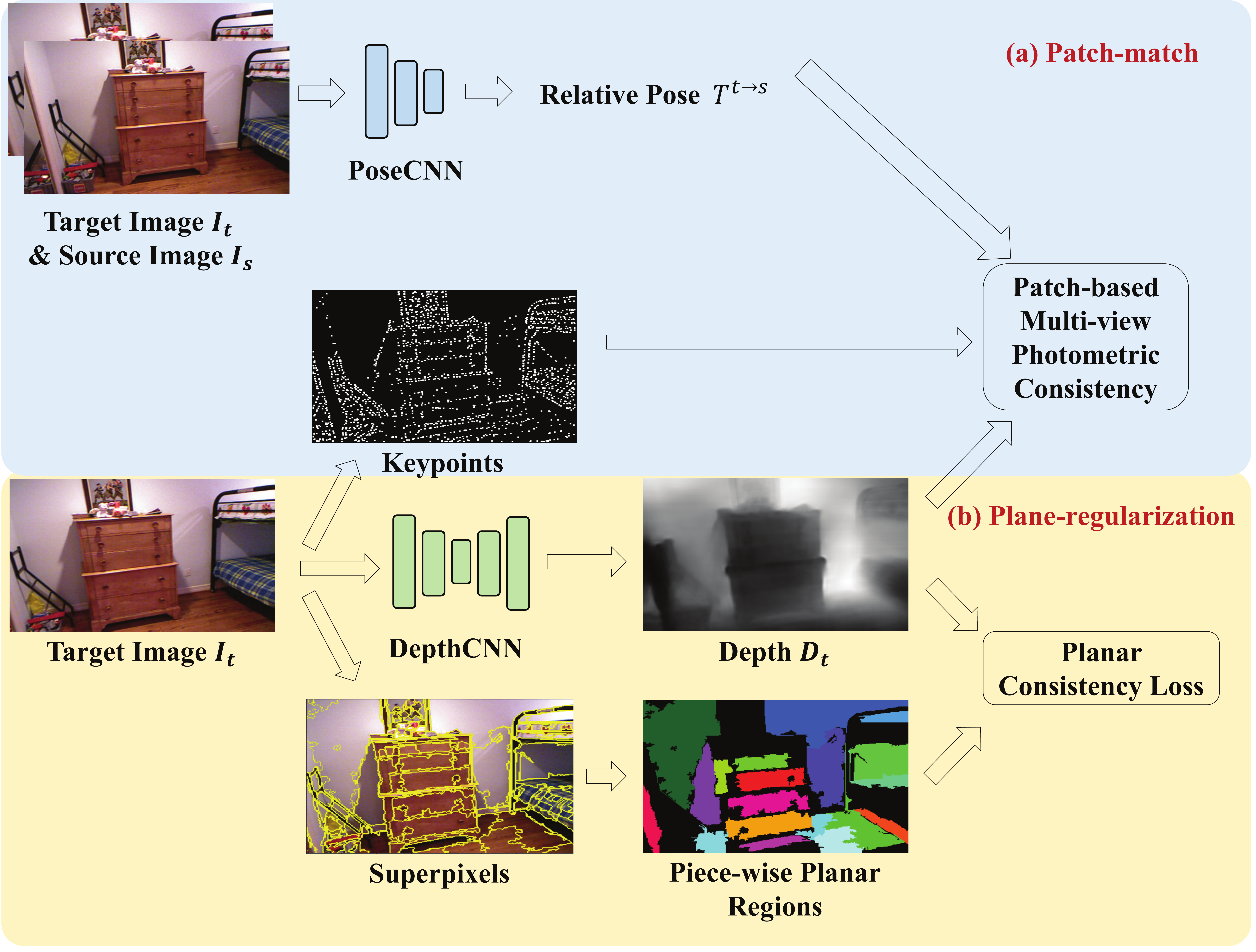}
\caption{Overall network architecture. Given input images, DepthCNN predicts the corresponding depth for the target image $I_t$, PoseCNN outputs the relative pose from the source to the target view. Our P$^2$Net consists of two parts: a) \textbf{Patch-match Module}: We warp the selected pixels along with their local neighbors with a patch-based warping module. b) \textbf{Plane-regularization Module}: We enforce depth consistency in large superpixel regions. }
\label{fig:network}
\end{figure}

\subsection{Overview}

Our goal is to learn a depth estimator for indoor environments with only monocular videos. Following recent success on unsupervised depth estimation~\cite{zhou2017unsupervised}, our P$^2$Net contains two learnable modules: DepthCNN and PoseCNN. DepthCNN takes a target view image $I_t$ as input and outputs its corresponding depth $D_t$. PoseCNN takes a source view image $I_s$ and a target view image $I_t$ as input and predicts the relative pose $T_{t \xrightarrow{} s}$ between two consecutive frames. A commonly used strategy is to first synthesize a novel view $I_t'$ with the predicted depth map $D_t$ and camera motion $T_{t \xrightarrow{} s}$, and minimize the photometric consistency error between the synthesized view $I_t'$ and its corresponding real view $I_t$. However, the training process soon collapses when directly applying this strategy to indoor scenarios. 

Our observation is that textured regions are beneficial to both depth estimation and camera motion estimation. In constrast, the large non-texture regions in indoor scenes might easily overwhelm the whole training process, and results are still blurred even these regions are masked out. Therefore, we propose to select representative keypoints that have large local variances. However, representing a point with a single intensity value, as done in previous unsupervised learning frameworks~\cite{godard2017unsupervised,godard2019digging}, is non-discriminative and may result in false matching. To address this problem, we propose a \textbf{Patch-match Module}, a patch-based representation that combines a point with the local window centered at that point to increase their discriminative abilities and minimize patch-based multi-view photometric consistency error. To handle the large non-texture regions, we propose a \textbf{Plane-regularization Module} to extract homogeneous-color regions using large superpixels and enforce that the predicted depth map within a superpixel may be approximated by a plane. The overview of our P$^2$Net is depicted in Fig.~\ref{fig:network}.

\subsection{Keypoints Extraction}
\begin{figure}
    \centering
    \includegraphics[height=5cm]{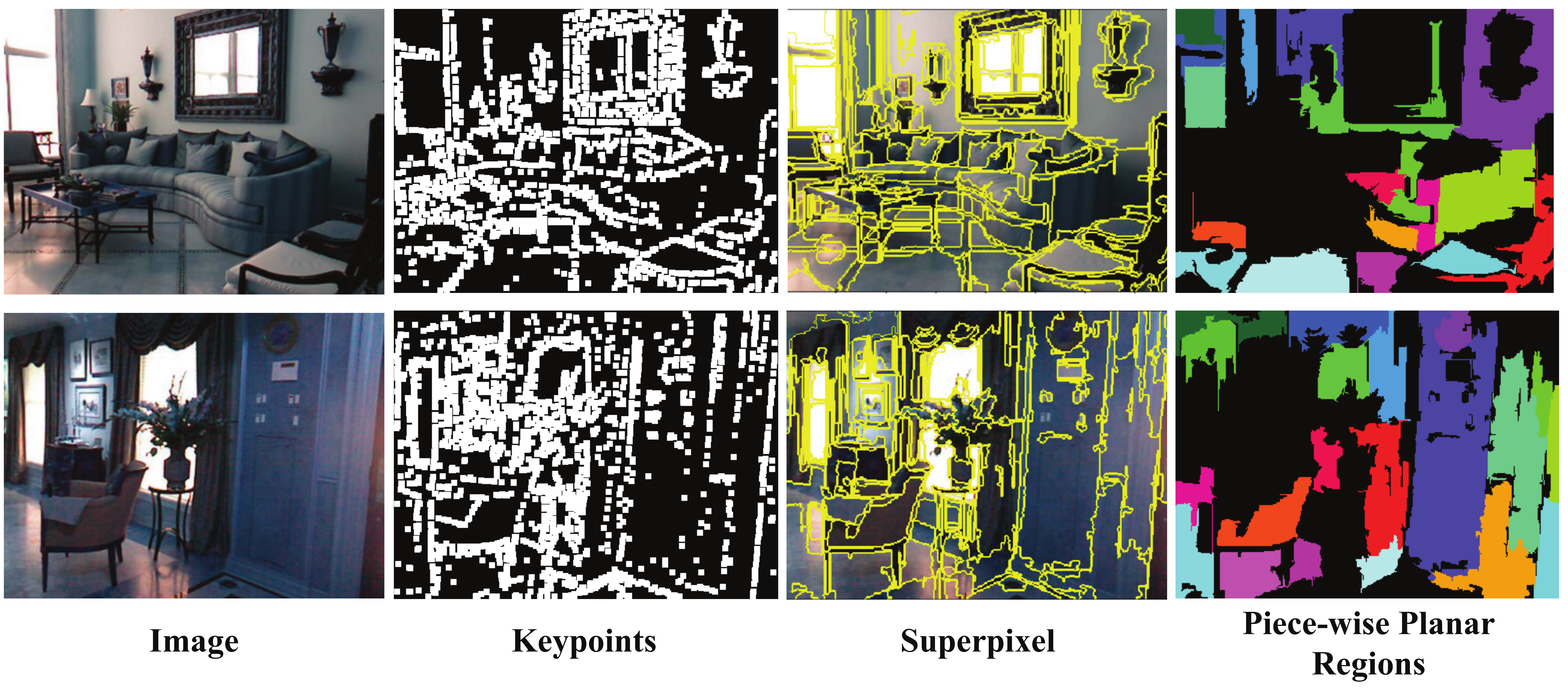}
    \caption{Examples of input images, their corresponding keypoints, superpixels and piece-wise planar regions obtained from large superpixels.}
    \label{fig:kp_spp}
\end{figure}

Different from outdoor scenes, the large proportion of the non-texture regions in indoor scenes can easily overwhelm the training process, leading to trivial solutions where DepthCNN always predicts an infinity depth, and PoseCNN always gives an identity rotation. Thus, only points within textured regions should be kept in the training process to avoid the network being stuck in such trivial results. Here, we adopt the points selection strategy from  
Direct Sparse Odometry (DSO)~\cite{engel2017direct} for its effectiveness and efficiency. Points from DSO are sampled from pixels that have large intensity gradients. 
Examples of extracted DSO keypoints are shown in Fig.~\ref{fig:kp_spp}. 

A critical benefit of our direct method over matching based approaches~\cite{zhou2019moving} is that we do not need to pre-compute the matching across images, which itself is a challenging problem. As a result, 
our points need to be extracted from the target image once only. No hand-crafted descriptor for matching is needed. Our method is hence more robust.
Also, note that our method is not limited to a specific type of keypoint detector. Other blob detectors, for example, SURF~\cite{bay2006surf}, also produce consistent results. 

\subsection{Patch-based Multi-view Photometric Consistency Error}
\begin{figure}
\centering
\includegraphics[height=3.3cm]{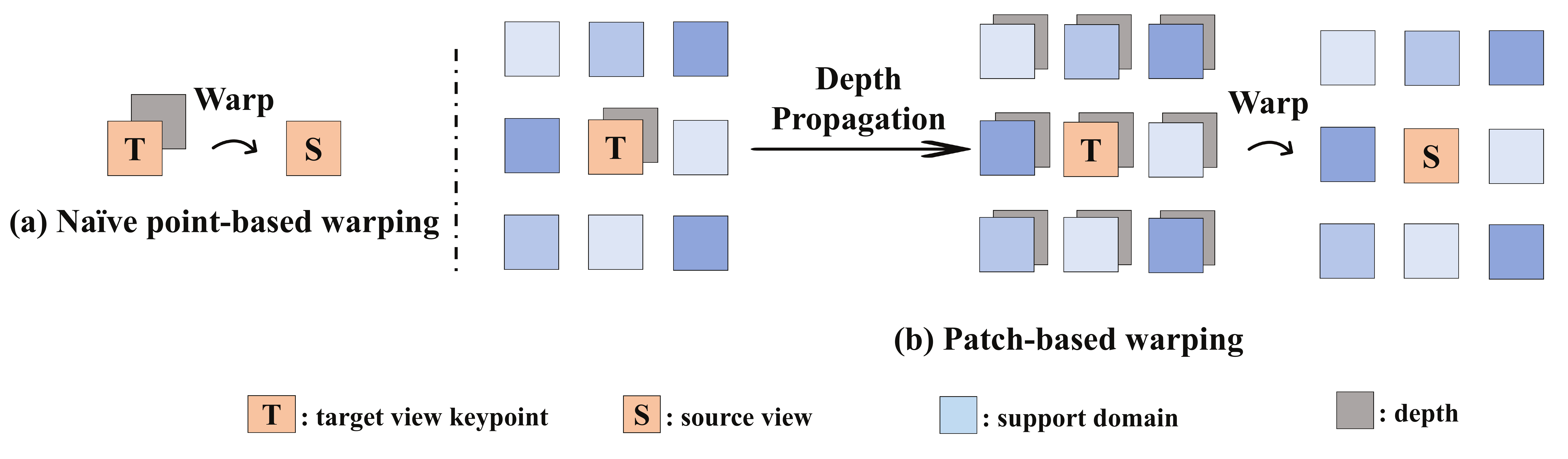}
\caption{Two types of warping operations. a) Naive point-based warping.  b) Our proposed patch-based warping. Note that we are defining pixels over its support domain and warp the entire window. Combining support domains into the pixel leads to more robust representations. Best viewed in color.}
\label{fig:patchwarp}
\end{figure}

With the extracted keypoints from the previous step, we can simply define a photometric consistency error by comparing the corresponding pixels' values. However, such point-based representation is not representative enough and may easily cause false matching because there are many pixels with the same intensity values in an image.
In traditional sparse SLAM pipelines~\cite{engel2017direct}, to overcome the above challenge, 
a support domain $\Omega_{p_i}$ is defined over each point $p_i$'s local window.  Photometric loss is then accumulated over each support domain $\Omega_{p_i}$ instead of a single isolated point. 
This operation would lead to more robust results as the extracted keypoints combined with their support domains are becoming much more unique.

Inspired from the above operation, here we propose a patch-based warping process as in Fig.~\ref{fig:patchwarp}.
Specifically, we extract DSO keypoints $p_{i}^t$ from the target view $t$, the original point-based warping process first back-projects the keypoints to the source view $I_s$ with:
\begin{equation}
    p_{i}^{t\xrightarrow{}s} = KT^{t\xrightarrow{}s}D(p_i)K^{-1}p_{i}^{t}
    \label{equ:flow}
\end{equation}
where $K$ denotes the camera intrinsic parameters, $T^{t\xrightarrow{}s}$ the relative pose between the source view $I_s$ and the target view $I_t$, and $D(p_i)$ the depth of point $p_i$.Then we sample the intensity values with bilinear interpolation~\cite{jaderberg2015spatial} at $p_{i}^{t\xrightarrow{}s}$ in the source view.

On the contrast, our approach assumes a same depth within each pixel's local window $\Omega_{p_i}^{t}$. Then, for every extracted keypoint, we warp the point together with its local support region $\Omega_{p_i}^{t}$ with the exact same depth.
Our warping process can thus be described as :
\begin{equation}
    \Omega_{p_i}^{t\xrightarrow{}s} = KT^{t\xrightarrow{}s}D(p_i)K^{-1}\Omega_{p_i}^{t}
    \label{equ:flow2}
\end{equation}
where $\Omega_{p_i}^{t}$ and $\Omega_{p_i}^{t \xrightarrow{}s}$ denotes the support domains of the point $p_i$ in the target view and the source view, respectively. 
From a SLAM perspective, we characterize each point over its support region, such patch-based approaches makes the representation of each point more distinctive and robust. 
From a deep learning perspective, our operation allows a larger region of valid gradients compared to the bilinear interpolation with only four nearest neighbors as in Equation (\ref{equ:flow}). 

Given a keypoint $p =(x,y)$, we define its support region $\Omega_p$ over a local window with size $N$ as:
\begin{equation}
\Omega_p = \{ (x+x_p, y+y_p), x_p \in \{-N, 0, N\}, y_p \in \{-N, 0, N\} \}
\end{equation}
$N$ is set to 3 in our experiments. 
Following recent work~\cite{godard2019digging}, we define our patch-based multi-view photometric consistency error as a combination of an L1 loss and a structure similarity loss SSIM~\cite{wang2004image} over the support region $\Omega_{p_i}$:
\begin{equation}
    L_{SSIM} = SSIM (I_t \left[\Omega_{p_i}^{t}\right],  I_s\left[\Omega_{p_i}^{t\xrightarrow{}s}\right] )
    \label{equ:flow2}
\end{equation}

\begin{equation}
    L_{L1} =|| I_t \left[\Omega_{p_i}^{t}\right] -  I_s\left[\Omega_{p_i}^{t\xrightarrow{}s}\right] ||_{1}
    \label{equ:flow2}
\end{equation}

\begin{equation}
    L_{ph} = \alpha L_{SSIM}  + (1 - \alpha) L_{L1}
\end{equation}
where $I_t\left[ p \right]$ denotes pixel values at $p$ in image $I_t$ via a bilinear interpolation, 
and $\alpha = 0.85$ a weighting factor. Note that when more than one source images are used in the photometric loss, we follow \cite{godard2019digging} to select the one with the minimum $L_{ph}$ for robustness purpose. We use a 3-frame (one target frame, 2 source frames) input in our ablation experiments and report the final results with a 5-frame (one target frame, 4 source frames) input.

\subsection{Planar Consistency Loss}
Finally, to further constrain the large non-texture regions in indoor scenes, we propose to enforce piecewise planar constraints into our network. Our assumption is that, most of the homogeneous-color regions are planar regions, and we can assume a continuous depth that satisfies the planar assumptions within these regions. Following representative work on reconstruction of indoor scenes~\cite{concha2015dpptam,concha2014using}, we adopt the Felzenszwalb superpixel segmentation~\cite{felzenszwalb2004efficient} in our approach. The segmentation algorithm follows a greedy approach and segments areas with low gradients, and hence produces more planar regions.
Examples with images, superpixels segmentation and piece-wise planar regions determined by superpixels, are demonstrated in Figure~\ref{fig:kp_spp}.
We can see that our assumption is reasonable, since indoor scenes generally consists of many man-made objects, like floor, walls, roof, \emph{etc}. Further, previous work also shows the good performance of indoor scene reconstruction with a piece-wise planar assumption in~\cite{liu2019planercnn,liu2018planenet,yu2019single}. 

Specifically, given an input image $I$, we first extract superpixels from the image and only keep regions larger than 1000 pixels. An intuition is that  the planar regions, like walls, floor, the surface of a table, are more likely to be within a larger area. 
Given an extracted superpixel $SPP_m$ and its corresponding depth $D(p_n)$ from an image,  where $p_n$ enumerates all the pixels within $SPP_m$, we first backproject all the points $p_n$ back to 3D space,
\begin{equation}
p_{n}^{3D} = D(p_n)K^{-1}p_n, p_n \subseteq SPP_m
\end{equation}
where $p_n^{3D}$ denotes the corresponding point of $p_n$ in 3D world. 
We define the plane in 3D following~\cite{liu2018planenet,yu2019single} as
\begin{equation}
A_{m}^{\top} p_{n}^{3D} = \mathbf{1} 
\end{equation}
where $A_{m}$ is plane parameter of $SPP_m$.

We use a least square method to fit the plane parameters $A_{m}$. Mathematically, we form two data matrices $Y_{m}$ and $P_n$, where $Y_{m} = \mathbf{1} = \begin{bmatrix} 1 & 1 & ... & 1 \end{bmatrix}^{\top}$, $P_{n} = \begin{bmatrix} p_1^{3D} & p_2^{3D} ... & p_n^{3D} \end{bmatrix}^{\top}$:
\begin{equation}
P_{n} A_{m} = Y_{m}
\end{equation}
Then $A_{m}$ can be computed with a closed-form solution:
\begin{equation}
    A_m = \left(P_n^\top P_n + \epsilon E \right)^{-1}P_n^\top Y_m.
\end{equation}
where $E$ is an identity matrix, and $\epsilon$ a small scalar for numerical stability. After obtaining the plane parameters, We can then retrieve our fitted planar depth for each pixel within the superpixel $SPP_m$ as 
$D^{'}(p_n) = (A_{m}^{\top}K^{-1}p_{n})^{-1}$. We then add another constraint to enforce a low plane-fitting error within each superpixel:
\begin{equation}
    L_{spp} = \sum_{m=1}^{M} \sum_{n=1}^{N} | D(p_n) - D^{'}(p_n) | 
\end{equation}
Here $M$ denotes the number of superpixels, and $N$ number of pixels in each superpixel.

\subsection{Loss Function}
We also adopt an edge-aware smoothness term $L_{sm}$ over the entire depth map as that in \cite{godard2017unsupervised,godard2019digging}:
\begin{eqnarray}
L_{sm} &=& \left | \partial_x d^*_t   \right | e^{-\left | \partial_x I_t \right |} + \left | \partial_y d^*_t   \right | e^{-\left | \partial_y I_t \right |},
\end{eqnarray} where $\partial_x$ denotes the gradients along the $x$ direction, $\partial_y$ along the $y$ direction and $d^*_t = d_t / \overline{d_t}$ is the normalized depth.

Our overall loss function is defined as :
\begin{equation}
    L=L_{ph}+\lambda_{1}L_{sm}+\lambda_{2}L_{spp}
\end{equation}
where $\lambda_{1}$ is set to $0.001$, $\lambda_{2}$ is set to 0.05 in our experiments.

\section{Experiments}
\label{sec:exp}
\subsection{Implementation Details}
We implement our solution under the PyTorch~\cite{paszke2019pytorch} framework.  
Following the pioneer work on unsupervised depth estimation in outdoor scenes, we use the same encoder-decoder architecture as that in~\cite{godard2019digging} with separate ResNet18s~\cite{he2016deep} pretrained on ImageNet as our backbones. 
We also adopt the same PoseCNN as that in~\cite{godard2019digging}, which takes only two frames as the input and output one pose. 
Adam~\cite{kingma2014adam} is adopted as our optimizer. The network is trained for a total of 41 epochs with a batch size of 12. Initial learning rate is set to $1e-4$ for the first 25 epochs. Then we decay it once by 0.1 for the next 10 epochs. We adopt random flipping and color augmentation during training. All images are resized to $288 \times 384$ pixels during training. Predicted depth are up-sampled back to the original resolution during testing. Since unsupervised monocular depth estimation exists scale ambiguity, we adopt the same median scaling strategy as that in \cite{godard2019digging,zhou2017unsupervised} for evaluation. A larger baseline is also beneficial for training, and we use a 5-frame input for the final result. For easy batch implementation, besides the standard DSO keypoints, we also draw points randomly to have a fixed number of 3K points from one image. 

\begin{table*}[!t] 
    \begin{center}
        \begin{tabular}{c|c|ccc|ccc}
            \toprule[1pt]
            Methods  & Supervised & rms $\downarrow$ &  rel $\downarrow$ &  log10 $\downarrow$ & $ \delta < 1.25$ $\uparrow$  & $ \delta < 1.25^2$ $\uparrow$  & $ \delta < 1.25^3$ $\uparrow$ \\
            \hline
            Make3D~\cite{saxena2008make3d} & \checkmark & 1.214 & 0.349 & - & 0.447 & 0.745 & 0.897 \\ 
            Liu et al.~\cite{liu2014discrete} & \checkmark & 1.200 & 0.350 & 0.131 & - & - & - \\
            Ladicky et al.~\cite{ladicky2014pulling} & \checkmark & 1.060 & 0.335 & 0.127 & - & - & - \\
            Li et al.~\cite{li2015depth} & \checkmark & 0.821 & 0.232 & 0.094 & 0.621 & 0.886 & 0.968 \\
            Liu et al.~\cite{liu2015learning} & \checkmark & 0.759 & 0.213 & 0.087 & 0.650 & 0.906 & 0.976 \\
            Li et al.~\cite{li2017two} & \checkmark & 0.635 & 0.143 & 0.063 & 0.788 & 0.958 & 0.991 \\
            Xu et al.~\cite{xu2017multi} & \checkmark & 0.586 & 0.121 & 0.052 & 0.811 & 0.954 & 0.987 \\
            DORN~\cite{fu2018deep} & \checkmark & 0.509 & 0.115 & 0.051 & 0.828 & 0.965 & 0.992 \\
            Hu et al.~\cite{hu2019revisiting} & \checkmark & 0.530 & 0.115 & 0.050 & 0.866 & 0.975 & 0.993 \\
            \hline
            PlaneNet~\cite{liu2018planenet} & \checkmark & 0.514 & 0.142 & 0.060 & 0.827 & 0.963 & 0.990 \\
            PlaneReg~\cite{yu2019single} & \checkmark & 0.503 & 0.134 & 0.057 & 0.827 & 0.963 & 0.990 \\
            
            \hline
            MovingIndoor~\cite{zhou2019moving} & $\times$ & 0.712 & 0.208 & 0.086 & 0.674 & 0.900 & 0.968 \\
            Monov2~\cite{godard2019digging} & $\times$ & 0.617 & 0.170 & 0.072 & 0.748 & 0.942 & 0.986 \\
            P$^2$Net (3 frames) & $\times$ & \textbf{0.599} & \textbf{0.159} & \textbf{0.068} & \textbf{0.772} & \textbf{0.942} & \textbf{0.984} \\
            \hline
            P$^2$Net (5 frames) & $\times$ & 0.561 & 0.150 & 0.064 & 0.796 & 0.948 & 0.986 \\
            P$^2$Net (5 frames PP)
            & $\times$ & \textbf{0.553} & \textbf{0.147} & \textbf{0.062} & \textbf{0.801} & \textbf{0.951} & \textbf{0.987} \\
            
            \hline 
            ResNet18 & \checkmark & 0.591 & 0.138 & 0.058 & 0.823 & 0.964 & 0.989 \\
            \toprule[1pt]
        \end{tabular}
    \end{center}
    \caption{Performance comparison on the NYUv2 dataset. We report results of depth supervised approaches in the first block, plane supervised results in the second block, unsupervised results in the third and fourth block, and the supervised upper bound of our approach denoted as ResNet18 in the final block. PP denotes the final result with left-right fliping augmentation in evaluation. Our approach achieves state-of-the-art performance among the unsupervised ones. $\downarrow$ indicates the lower the better, $\uparrow$ indicates the higher the better.}
    \label{tab:nyuv2}
\end{table*}
\subsection{Datasets}
We evaluate our P$^2$Net on two publicly available datasets of indoor scenes, including NYU Depth V2~\cite{nyu} and ScanNet~\cite{dai2017scannet}. 

\textbf{NYU Depth V2.} NYU Depth V2 is captured with a Microsoft Kinect sensor and consists of a total 582 indoor scenes. We adopt the same train split of 283 scenes following previous work on indoor depth estimation~\cite{zhou2019moving} and provide our results on the official test set with the standard depth evaluation criteria. We sample the training set at 10 frames interval as our target views and use $\pm 10$, $\pm 20$ frames as our source views. This leaves us around 20K unique images, a number much less than the 180K images used in the previous work of unsupervised indoor depth estimation~\cite{zhou2019moving}. Training takes around 15 hours on one P40 GPU. Note that the original NYU Depth V2 images are unaligned. We undistort the input image as in~\cite{teed2018deepv2d} and crop 16 black pixels from the border region.  
 
\begin{figure*}[!htbp]
    \centering
    \includegraphics[height=16cm]{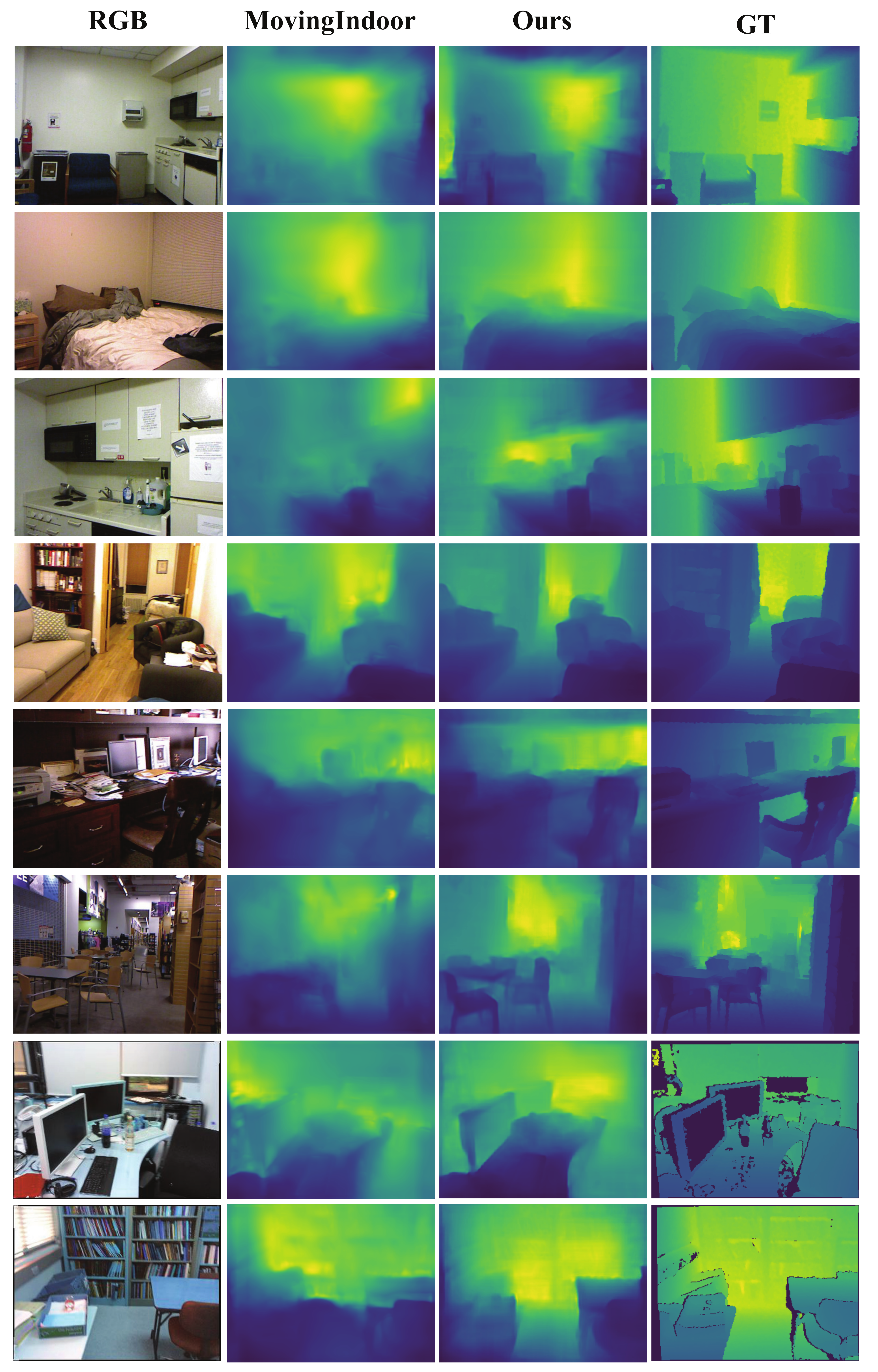}
    \caption{Depth visualization on NYUv2 (first 6 rows) and ScanNet (last 2 rows). We trained our model on NYUv2 and directly transfer the weights to ScanNet without fine-tunning. From left right: input image, results of MovingIndoor~\cite{zhou2019moving}, our results and ground truth depth. GeoNet would collapse on indoor datasets due to the large non-texture regions. Compared to MovingIndoor~\cite{zhou2019moving}, our methods preserve more details. }
    \label{fig:compare_nyuv2_scannet}
\end{figure*}

Quantitative results are provided in Table~\ref{tab:nyuv2}. We compare with MovingIndoor~\cite{zhou2019moving}, the pioneer work on unsupervised indoor depth estimation and Monov2~\cite{godard2019digging}, a state-of-the-art unsupervised depth estimation method on outdoor datasets. Note that our proposed single-scale method is even able to achieve superior performance even when compared to multi-scale approaches like \cite{godard2019digging}. We further provide some visualization of our predicted depth in Fig.~\ref{fig:compare_nyuv2_scannet}. GeoNet collapsed during training as we inspected. Compared to MovingIndoor~\cite{zhou2019moving}, our method preserves much more details owing to the patch-based multi-view consistency module. A supervised upper bound, denoted as ResNet18, is also provided here by replacing the backbone network in~\cite{hu2019revisiting} with ours.

\begin{figure}[!ht]
    \label{fig:ablation}
    \centering
    \includegraphics[height=8cm]{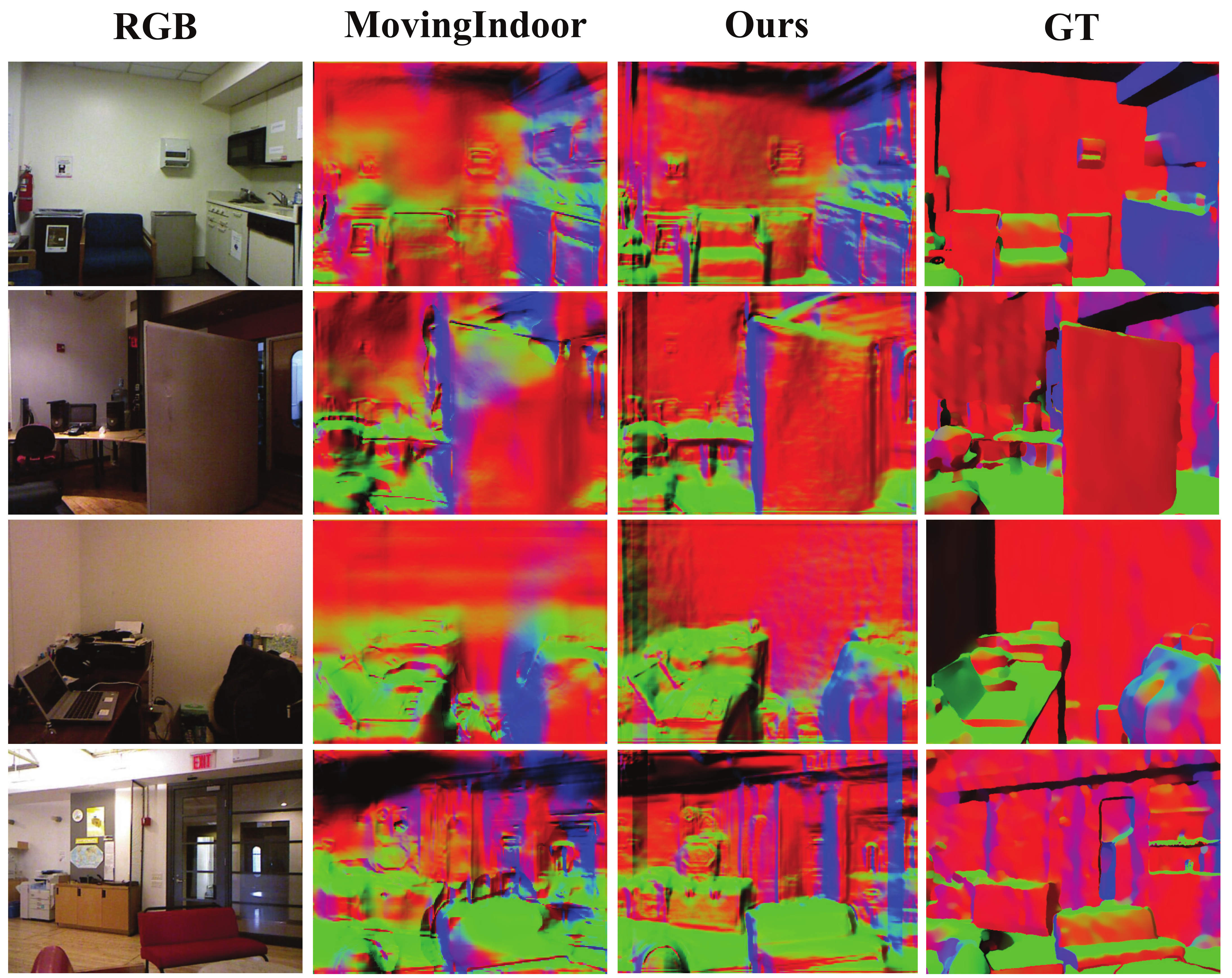}
    \caption{Visualization of fitted surface norm from 3D point clouds on the NYUv2 dataset. From left to right: input image, results of MovingIndoor~\cite{zhou2019moving}, ours and ground truth normal. Our method produces more smooth results in planar regions.}
    \label{fig:compare_nyuv2_norm}
\end{figure}

We also provide our results for surface normal estimation in Tab.~\ref{tab:nyu_norm}. Surface normal in our method is fitted directly from the point clouds within a local window. Not only is our result the best among the unsupervised ones, it is also close to supervised results like DORN~\cite{fu2018deep}. We visualize some results of our method for surface normal estimation in Fig.~\ref{fig:compare_nyuv2_norm}.

\textbf{Scannet.} Scannet~\cite{dai2017scannet} is captured with Structure sensor attached to a handheld device, containing around 2.5M images captured in 1513 scenes. While there is no current official train/test split on ScanNet for depth estimation, we randomlly pick 533 testing images from diverse scenes. We directly evaluate our models pretrained on NYUv2 under a transfer learning setting to test the generalizability of our approach. We showcase some of the prediction results in Fig.~\ref{fig:compare_nyuv2_scannet}. We achiever better result as reported in Tab.~\ref{tab:transfer}. 

\begin{table}[!ht]
\centering
\begin{tabular}{c|c|c|cccc}
\toprule[1pt]
Methods & Supervised & Mean $\downarrow$ & $11.2\degree \uparrow$ & $22.5\degree \uparrow$  & $30\degree \uparrow$ \\
           \hline \hline
\multicolumn{6}{c}{Predicted Surface Normal from the Network}               \\ \hline
3DP \cite{fouhey2013data}  & \checkmark &  33.0   & 18.8  & 40.7   & 52.4    \\
Ladicky \etal.\ \cite{zeisl2014discriminatively}  & \checkmark & 35.5   & 24.0  & 45.6  & 55.9   \\
Fouhey \etal.\ \cite{fouhey2014unfolding}  & \checkmark &  35.2  & 40.5 & 54.1 & 58.9    \\
Wang \etal.\ \cite{wang2015designing}  & \checkmark &   28.8   & 35.2 & 57.1 & 65.5  \\
Eigen \etal.\ \cite{eigen2015predicting}  & \checkmark& 23.7  & 39.2 & 62.0 & 71.1    
\\\hline
\multicolumn{6}{c}{Surface Normal Fitted from Point Clouds}               \\ \hline
GeoNet~\cite{qi2018geonet} & \checkmark & 36.8 & 15.0  & 34.5 & 46.7 \\
DORN~\cite{fu2018deep}  & \checkmark & 36.6 & 15.7   & 36.5   & 49.4 \\
MovingIndoor~\cite{zhou2019moving} & $\times$ & 43.5 & 10.2 & 26.8 & 37.9 \\
Monov2~\cite{godard2019digging} & $\times$ & 43.8 & 10.4 & 26.8 & 37.3    \\ \hline
P$^2$Net (3 frames)& $\times$ & 38.8 & 11.5 & 31.8 & 44.8     \\
P$^2$Net (5 frames)& $\times$ & 36.6 & 15.0 & 36.7 & 49.0     \\
P$^2$Net (5 frames pp)& $\times$ & \textbf{36.1} & \textbf{15.6} & \textbf{37.7} & \textbf{50.0}     \\
\toprule[1pt]
\end{tabular}
\caption{Surface normal evaluation on NYUv2. We report the results of methods that directly predict surface normal from the network in the first block. Results that are fitted from the point cloud are provided in the second and the third block. PP denotes the final result with left-right fliping augmentation in evaluation. The performance of our method is even close to some of the supervised approaches. }
\label{tab:nyu_norm}
\end{table}

\begin{table*}[ht] 
\begin{center}
\begin{tabular}{c|ccc|ccc}
  \toprule[1pt]
  Methods  & rms $\downarrow$ &  rel $\downarrow$ &  log10 $\downarrow$ & $ \delta < 1.25$ $\uparrow$  & $ \delta < 1.25^2$ $\uparrow$  & $ \delta < 1.25^3$ $\uparrow$ \\ \hline
  MovingIndoor~\cite{zhou2019moving} & 0.483 & 0.212 & 0.088 & 0.650 & 0.905 & 0.976 \\
  Monov2~\cite{godard2019digging} & 0.458 & 0.200 & 0.083 & 0.672 & 0.922 & 0.981 \\
  P$^2$Net & \textbf{0.420} & \textbf{0.175} & \textbf{0.074} & \textbf{0.740} & \textbf{0.932} & \textbf{0.982} \\
  \toprule[1pt]
\end{tabular}
\end{center}

\caption{Performance comparison on transfer learning. Results are evaluated directly with NYUv2 pretrained models on ScanNet. Our model still achieves the best result.}
\label{tab:transfer}
\end{table*}

\subsection{Ablation Experiments}
\textbf{The effect of Patch-match and Plane-regularization.}
 For our baseline, we first calculate the variance within a local region for each pixel. This servers as our texture/non-texture region map. Photometric loss is directly multiplied by the map. This represents the most straightforward case when only point-based supervision is provided. We report the numbers in the first row of Tab.~\ref{tab:ablation}. Then we add our proposed Patch-match module and report the results in the second line, the Plane-regularization module in the fourth line. Experiments demonstrate the effectiveness of our proposed modules.
 
 \textbf{Different keypoint types.} Here, we demonstrate that our method is not limited to some specific type of keypoint detectors. We replace DSO with a blob region detector SURF~\cite{bay2006surf}. We achieve similar results as reported in line two and three in Tab.~\ref{tab:ablation}.
 
 \begin{table}[]
     \centering
     \begin{tabular}{c|c|c|cc|ccc}
         \toprule[1pt]
         Keypoint & \begin{tabular}[c]{@{}c@{}}Patch \\ Match\end{tabular} & \begin{tabular}[c]{@{}c@{}}Plane \\ Regularization\end{tabular} & rms $\downarrow$ &  rel $\downarrow$ &  $ \delta < 1.25$ $\uparrow$  & $ \delta < 1.25^2$ $\uparrow$  & $ \delta < 1.25^3$ $\uparrow$ \\ \hline
         -  & & & 0.786 & 0.240 & 0.628& 0.884 & 0.962   \\
         DSO & \checkmark & & 0.612 & 0.166 & 0.758 & 0.945 & 0.985 \\
         SURF & \checkmark & & 0.622 & 0.169 & 0.750 & 0.941 & 0.986  \\
         DSO  & \checkmark & \checkmark &  \textbf{0.599} & \textbf{0.159} &  \textbf{0.772}  &   \textbf{0.942}  &   \textbf{0.984}    \\ \hline
         \toprule[1pt]
     \end{tabular}
     \caption{Ablation study of our proposed module on the NYUv2 dataset.}
 \label{tab:ablation}
 \end{table}

\textbf{Camera pose.} Following previous work~\cite{teed2018deepv2d} on predicting depth from videos, we provide our camera pose estimation results on the ScanNet dataset, consisting a total of 2000 pairs of images from diverse scenes. Note that since our method is monocular, there exists scale ambiguity in our predictions. Hence, we follow~\cite{teed2018deepv2d} and rescale our translation during evaluation. 
Results are reported in Tab.~\ref{tab:pose}. Our method performs better than MovingIndoor~\cite{zhou2019moving}.

\textbf{Results on outdoor scenes.} Here we also provide our results on the KITTI benchmark in Tab.~\ref{tab:kitti}. We trained and evaluated our results on the same subset as in ~\cite{godard2019digging}. Our method is compared against our baseline, Monov2~\cite{godard2019digging} and MovingIndoor~\cite{zhou2019moving}. Our method outperforms another unsupervised indoor depth estimation approach MovingIndoor. Please note that different from indoor scenes, the main challenge in outdoor scenes are moving objects (like cars) and occlusions, which seldom occur in indoor scenes. Our method does not take such priors into consideration. On the contrast, Monov2 is specially designed to handle these cases.

\begin{table}[!htbp]
    \begin{minipage}{0.48\linewidth}
    \centering
    \begin{tabular}{c|ccc}
        \toprule[1pt]
        Method       & rot(deg) & tr(deg) & tr(cm) \\ \hline
        MovingIndoor~\cite{zhou2019moving} & 1.96     & 39.17   & 1.40   \\
        Monov2~\cite{godard2019digging}       & 2.03     & 41.12   & \textbf{0.83} \\
        P$^2$Net        & \textbf{1.86}     & \textbf{35.11}   & 0.89   \\ \hline
       \toprule[1pt]
    \end{tabular}
    \caption{Camera pose estimation results.}
    \label{tab:pose}
    \end{minipage}
    \begin{minipage}{0.48\linewidth}  
        \centering
    \begin{tabular}{c|ccc} \toprule[1pt]
        Method & rel$\downarrow$ & rms $\downarrow$ &$ \delta < 1.25$ $\uparrow$ \\ \hline
        MovingIndoor~\cite{zhou2017unsupervised} & 0.130    & 5.294   & -      \\
        P$^2$Net      & \textbf{0.126} & \textbf{5.140}  & \textbf{0.862}  \\ \hline
        Monov2~\cite{godard2019digging}    & 0.115    & 4.863   & 0.877  \\ \hline\toprule[1pt]
    \end{tabular}
    \caption{Results on KITTI.}
    \label{tab:kitti}
    \end{minipage}
\end{table}

\section{Conclusion}
This paper addresses the challenging unsupervised depth estimation task in indoor scenes with large areas of non-texture regions.  
We propose P$^2$Net that uses patches centered at discriminative points as their representations and warp patches instead of points, and use superpixels to represent each plane and enforce a low plane-fitting error.
Extensive experiments on NYUv2 and ScanNet validate the effectiveness of our P$^2$Net. 
Here for simplicity we adopt the fronto-parallel assumption. One possible solution could be to first pretrain the network and calculate normal from depth. Then we can combine normal into the training process.

\section*{Acknowledgements}
The work was supported by National Key R\&D Program of China (2018AAA0100704), NSFC \#61932020,
and ShanghaiTech-Megavii Joint Lab. We would also like to thank Junsheng Zhou from Tsinghua University for detailed commonts of reproducing his work and some helpful discussions.

\clearpage
%
%
\appendix
\begin{appendix}
\section{Surface normal visualization}
We provide more visualizations of surface normal prediction on the ScanNet~\cite{dai2017scannet} dataset. In our implementation, we directly fit the surface normal from ground truth depth annotation. Black pixels indicate invalid regions where no ground truth depths are provided. Compared to MovingIndoor~\cite{zhou2019moving}, our surface normal estimation better preserves the boundary of the planar regions, thanks to our superpixel constraint. 

\begin{figure}[!t]
    \centering
    \includegraphics[height=20cm]{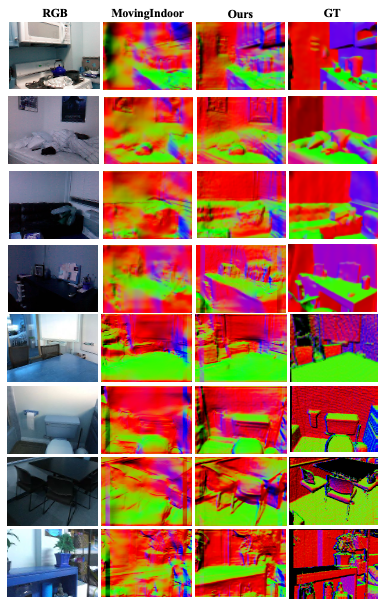}
    \caption{Visualization of surface normal results on the Scannet~\cite{dai2017scannet} dataset. From left to right: input RGB, MovingIndoor~\cite{zhou2019moving}, our results and surface normal fitted from ground truth depth. Black pixels in ground truth indicate invalid regions where no depth ground truth are provided.}
    \label{fig:scannet_norm}
\end{figure}

\section{Point cloud visualization}
\begin{figure}[!htbp]
    \centering
    \includegraphics[height=13.5cm]{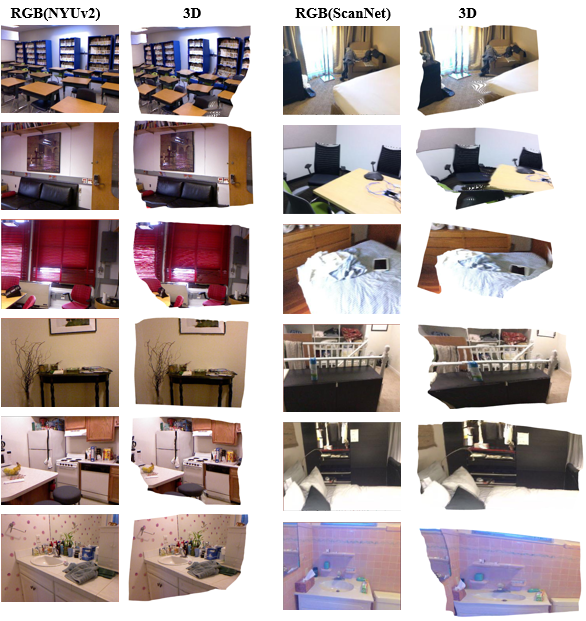}
    \caption{Point cloud visualization. From left to right: input RGB from NYUv2, point cloud in 3D, RGB from ScanNet, point cloud in 3D.}
    \label{fig:pointcloud}
\end{figure}

We further provide some point cloud visualization on NYUv2~\cite{nyu} and ScanNet~\cite{dai2017scannet} dataset in Figure~\ref{fig:pointcloud}.

\section{The effect of different patterns.} 
Here, we compare the effect of different patterns in our Patch-match module. We experiment with different $N$s and report the result in Table~\ref{tab:diffpattern}. Setting $N$ to 3 gives best results. On the contrast, keeping a larger pattern ($N=4$) might introduce additional noise, this would lead to a decay in performance.
\begin{table}[!h] 
    \begin{center}
        \begin{tabular}{c|cc|ccc}
            \toprule[1pt]
            $N$  & rms $\downarrow$ &  rel $\downarrow$ & $ \delta < 1.25$ $\uparrow$  & $ \delta < 1.25^2$ $\uparrow$  & $ \delta < 1.25^3$ $\uparrow$ \\ \hline
            1 & 0.629 & 0.173 & 0.746 & 0.939 & 0.984  \\
            2 & 0.618 & 0.170 & 0.748 & 0.937 & 0.984 \\
            3 & \textbf{0.612} & \textbf{0.166} & \textbf{0.758} & \textbf{0.945} & \textbf{0.985} \\
            4 & 0.634 & 0.173 & 0.741 & 0.938 & 0.984 \\
            \toprule[1pt]
        \end{tabular}
    \end{center}
    \caption{Comparison between different patterns in our Patch-match module.}
    \label{tab:diffpattern}
\end{table}
\end{appendix}
\clearpage

\bibliographystyle{splncs04}
\bibliography{egbib}

\end{document}